\pdfoutput=1

\documentclass[journal]{IEEEtran}
\ifCLASSINFOpdf
\else
\fi
\usepackage{times}
\usepackage{url}
\usepackage{latexsym}
\usepackage{xcolor}
\usepackage{graphicx}
\usepackage{amsmath}
\usepackage{subfigure}
\usepackage{multirow}
\usepackage{booktabs}
\usepackage{microtype}
\usepackage{stfloats}
\usepackage{amsfonts,amssymb}
\usepackage{bbm}
\usepackage{hyperref}
\usepackage{underscore}
\usepackage{xcolor}
\usepackage{cancel}
\usepackage{algorithm}
\usepackage{algpseudocode}
\usepackage{amsmath}
\usepackage{algpseudocode}
\usepackage{mathrsfs}
 
\usepackage[numbers,sort&compress]{natbib}

\begin{document}
%
\title{PiSLTRc: Position-informed Sign Language Transformer with Content-aware Convolution}
%
%
%

\author{Pan Xie, Mengyi Zhao, Xiaohui Hu
\thanks{Pan Xie and Mengyi Zhao are with the School of Automation Science and Electrical Engineering, Beihang University, Beijing 100191, China (e-mail:
panxie@gmail.com, mengyizhao@buaa.edu.cn).}
\thanks{Xiaohui Hu is with the Science and Technology on Integrated Information System Laboratory, Institute of Software, Chinese Academy of Sciences, Beijing 100191, China (e-mail: hxh@iscas.ac.cn)}}
\maketitle

\begin{abstract}
Since the superiority of Transformer in learning long-term dependency, the sign language Transformer model achieves remarkable progress in Sign Language Recognition (SLR) and Translation (SLT). However, there are several issues with the Transformer that prevent it from better sign language understanding. The first issue is that the self-attention mechanism learns sign video representation in a frame-wise manner, neglecting the temporal semantic structure of sign gestures. Secondly, the attention mechanism with absolute position encoding is direction and distance unaware, thus limiting its ability. To address these issues, we propose a new model architecture, namely PiSLTRc, with two distinctive characteristics: (i) content-aware and position-aware convolution layers. Specifically, we explicitly select relevant features using a novel content-aware neighborhood gathering method. Then we aggregate these features with position-informed temporal convolution layers, thus generating robust neighborhood-enhanced sign representation. (ii) injecting the relative position information to the attention mechanism in the encoder, decoder, and even encoder-decoder cross attention. Compared with the vanilla Transformer model, our model performs consistently better on three large-scale sign language benchmarks: PHOENIX-2014, PHOENIX-2014-T and CSL. Furthermore, extensive experiments demonstrate that the proposed method achieves state-of-the-art performance on translation quality with $+1.6$ BLEU improvements.
\end{abstract}


%
\IEEEpeerreviewmaketitle

\section{Introduction}\label{Introduction}
\IEEEPARstart{S}{ign} language (SL) is a native language of people with disabled hearing. As a visual language, it consists of various hand gestures, movements, facial expressions, transitions, etc. Sign Language Recognition (SLR) and Translation (SLT) aim at converting the video-based sign languages into sign gloss sequences and spoken language sentences, respectively. Most previous works in this field focus on continuous SLR with the gloss supervision~\cite{Koller2016DeepHH,Koller2020WeaklySL,Cui2017RecurrentCN,Pu2018DilatedCN,Wang2019ANS,Cui2019ADN,Pu2019IterativeAN,Pu2020BoostingCS,Hu2020GloballocalEN,Cheng2020FullyCN,Camgz2017SubUNetsEH,zhou2020spatial,8466903}, few attempts have been made for SLT~\cite{Camgz2018NeuralSL,Camgz2020SignLT,Li2020TSPNetHF,Camgz2020MultichannelTF}. The main difference is that gloss labels are in the same order with sign gestures, and thus the gloss annotations significantly ease the syntactic alignment under the SLR methods. However, the word ordering rules in natural language are distinct from their counterparts in video-based sign languages~\cite{Pfau2018TheSO}. Moreover, sign videos are composed of continuous sign gestures represented by sub-video clips without explicit boundaries. Therefore, directly learning the mapping between frame-wise signs and natural language words is challenging. 

To achieve better translation performance, a promising research line is to perform joint sign language recognition and translation model, which recognizing glosses and translating natural language sentences simultaneously~\cite{Camgz2018NeuralSL, Camgz2020SignLT}. By doing so, learning with the glosses supervision can better understand sign videos and bring significant benefits to sign language translation. Along this line, Camgzet \textit{et al.}~\cite{Camgz2020SignLT} proposes a joint model, Sign Language Transformer (SLTR), which is based on vanilla Transformer~\cite{Vaswani2017AttentionIA}. They learn recognition and translation simultaneously and achieve state-of-the-art results due to the Transformer's advantage in sequence modeling tasks. However, there are still some inherent flaws that limit the capabilities of the Transformer model when solving the SLR and SLT tasks:
\begin{itemize}
\item[(a)] The self-attention mechanism aggregates temporal sign visual features in a frame-wise manner. This mechanism neglects the temporal structure of sign gestures represented by sub-videos, leading to substantial ambiguity in recognition and translation.
\item[(b)] The attention mechanism is permutation-insensitive. Thus position encoding is essential to inject position information for sequence learning, e.g., sign video learning and sentence learning. However, the absolute position encoding used in vanilla Sign Language Transformer (SLTR)~\cite{Camgz2020SignLT} is demonstrated distance and direction unaware~\cite{Shaw2018SelfAttentionWR,Yan2019TENERAT}, thus limit its ability for better performance.
\end{itemize}

To remedy this first shortcoming (a), an intuitive idea is to gather neighboring temporal features to enhance the frame-wise sign representation. However, it is difficult to determine the boundaries of a sign gesture and select the surrounding neighbors precisely. In this paper, we propose a Content-aware and Position-aware Temporal Convolution (CPTcn) to learn robust sign representations. We first propose a content-aware neighborhood gathering method to adaptively select the surrounding neighbors. Specifically, we leverage the local consistency of sign gestures. That is to say, adjacent frames that  belong to a sign gesture share similar semantics. Accordingly, we dynamically select neighboring features based on the similarities. Then we aggregate the selected features with temporal convolution layers. However, temporal convolution with a limited receptive field is insufficient to capture the position information of the features in the selected region~\cite{Islam2020HowMP}. To alleviate the drawback, we inject position awareness into convolution layers with Relative Position Encoding (RPE). By aggregating with neighboring similar features, our CPTcn module obtains discriminative sign representations, thus improving the recognition and translation results.

To solve the second issue (b), we inject relative position information into the learning of sign videos and target sentences. Furthermore, we consider the relative position between sign frames and target words. To the best of our knowledge, we are the first trying to model the position relationship between source sequence and target sequence in sequence-to-sequence architectures. There are several existing methods to endow the self-attention mechanism with relative position information~\cite{Shaw2018SelfAttentionWR,Dai2019TransformerXLAL,He2020DeBERTaDB,Raffel2020ExploringTL,Ke2020RethinkingPE}. In this paper, we adopt the Disentangled Relative Position Encoding (DRPE)~\cite{He2020DeBERTaDB} in our video-based sign language learning, target sentence learning, and their mapping learning. Note that, different from RPE mentioned above, DRPE contains the correlations between relative position and sign features, which is proven effective to bring improvements~\cite{He2020DeBERTaDB,Yang2019XLNetGA}. With the distance and direction awareness learning from DRPE, our improved Transformer model learns better feature representations, thus gaining significant improvements. 

\begin{figure}[t]
\centering
\includegraphics[width=0.48\textwidth,height=0.33\textheight]{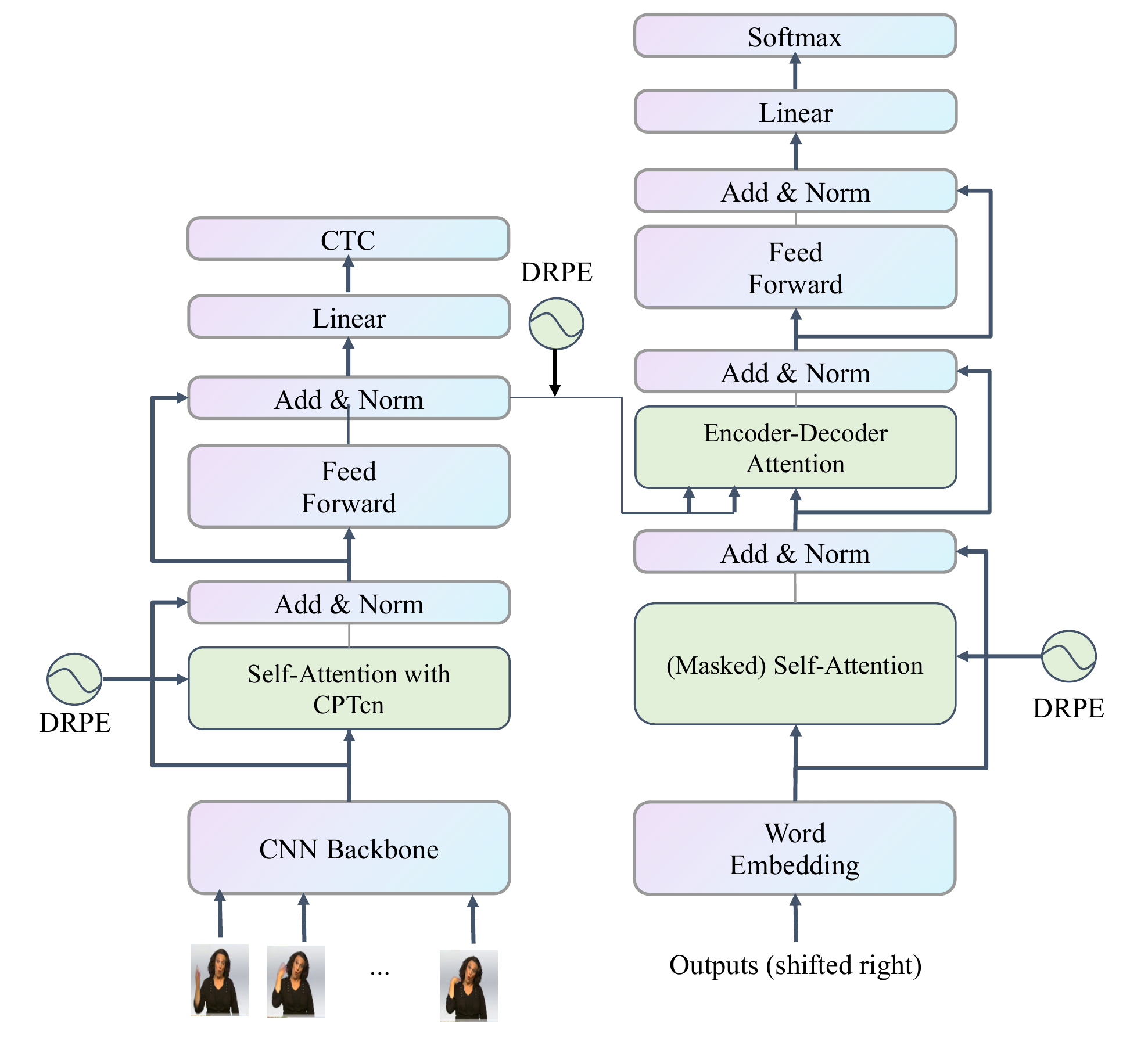}
\caption{\textbf{The overview of our sign language Transformer model equipped with Content-aware and Position-aware Temporal Convolution (CPTcn) and Disentangled Relative Position Encoding (DRPE).}}
\label{architecture} 
\end{figure}

We call our approach PiSLTRc for "\textbf{P}osition-\textbf{i}nformed \textbf{S}ign \textbf{L}anguage \textbf{TR}ansformer with content-aware \textbf{c}onvolution". The overview of our model can be seen in Figure~\ref{architecture}. The main technical contributions of our work are summarized as follows:

\begin{itemize}

\item [1)] We propose a content-aware and position-aware CPTcn module to learn neighborhood-enhanced sign features. Specifically, We first introduce a novel neighborhood gathering method based on the semantic similarities. Then we aggregate the selected features with position-informed temporal convolution layers. 

\item[2)] We endow the Transformer model with relative position information. Compared with absolute position encoding, relative position encoding performs better for sign video and natural sentence learning. Furthermore, we are the first to consider the relative position relationship between sign frames and target words. 

\item[3)] Equipped with the proposed two techniques, our model achieves state-of-the-art performance in translation accuracy on the largest public dataset RWTH-PHOENIX-Weather 2014\textbf{T}. Also, we obtain significant improvements in recognition accuracy compared with other RGB-based models on both PHOENIX-2014 and PHOENIX-2014-T dataset.

\end{itemize}

The remainder of this paper is organized as follows. Section~\ref{Related Work} reviews related works in sign language and position encoding. Section~\ref{Method} introduces the architecture of our proposed PiSLTRc model. Section~\ref{Experiments} provides implementation details on our model, presents a quantitative analysis that provides some intuition as to why our proposed techniques work, and finally presents the experimental results compared with several baseline models.

%
%
%
%

 

\section{Related Work} \label{Related Work}

\subsection{Sign Language Recognition}
Most previous sign language works focus on continuous sign language recognition (cSLR), which is a weakly supervised sequence labeling problem~\cite{Koller2020WeaklySL}. cSLR aims at transcribing video-based sign language into gloss sequence. With the released of larger-scale cSLR datasets ~\cite{Forster2014ExtensionsOT}, numerous researches burst out implementing sign language recognition tasks in an end-to-end manner~\cite{Koller2016DeepHH,Koller2020WeaklySL,Cui2017RecurrentCN,Cui2019ADN,Pu2018DilatedCN,Wang2019ANS,Pu2019IterativeAN,Pu2020BoostingCS,Hu2020GloballocalEN,Cheng2020FullyCN,Camgz2017SubUNetsEH,zhou2020spatial,8466903}. The gloss annotations are in same order with sign language, this monotonic relationship significantly ease the syntactic alignment with the cSLR methods. However, the relationship between gloss sequences and the spoken natural language is non-monotonic. Thus it is infeasible to realize SLT with cSLR methods. Fortunately, the knowledge learned by cSLR can be transferred to SLT models and facilitate their performance. 

\subsection{Sign Language Translation}
Sign language Translation (SLT) is much more challenging because the alignment learning of frame-wise sign gestures and natural language words is difficult. Camgz \textit{et al.}~\cite{Camgz2018NeuralSL} first introduce an end-to-end SLT model that uses Convolution Neural Networks (CNNs) backbone to capture spatial feature and utilizes attention-based encoder-decoder model~\cite{Luong2015EffectiveAT} to learn the mapping of sign videos and natural language sentences. Based on this work, Camgz \textit{et al.}~\cite{Camgz2020SignLT} replace the sequence-to-sequence structure with Transformer architecture~\cite{Vaswani2017AttentionIA} which is the state-to-the-art model in Neural Machine Translation (NMT) area. Furthermore, they jointly learn the sign language recognition and translation with a shared Transformer encoder and demonstrate that joint training provides significant benefits. Our work is built upon their joint sign language Transformer model, where we improve the Transformer with our proposed CPTcn module and endow the Transformer model with relative position information.

\subsection{Position Encoding in Convolution}
Temporal convolution neural network is a common method to model sequential information~\cite{9053841,8777194,8361904,8085174}. Convolution layer is demonstrated implicitly to learn absolute position information from the commonly used padding operation~\cite{Islam2020HowMP}. However, it is insufficient to learn powerful representations that encode sequential information, especially with the limited receptive field.  Explicitly encoding absolute position information is shown effective to learn image features~\cite{Islam2020HowMP}. Upon their hypothesis, we apply relative position encoding (RPE) to the temporal convolution layers, aiming to model the positional correlations between the current feature and its surrounding neighbors.

\subsection{Position Encoding in Self-attention}
Transformer entirely relies on the attention mechanism, which does not explicitly model the position information. To remedy the drawback, the sinusoidal absolute position encoding~\cite{Vaswani2017AttentionIA} and learnable absolute position encoding~\cite{Devlin2019BERTPO} are proposed to endow their model with position information. Afterward, relative position encoding is proposed to model long sequence~\cite{Yang2019XLNetGA} and provides the model with relation awareness~\cite{Yan2019TENERAT,Shaw2018SelfAttentionWR}. In our work, we reuse the disentangled position encoding~\cite{He2020DeBERTaDB} to exploit the distance and direction awareness with relative position encoding. Moreover, we also explore the position relationship between sign video and target sentence. Note that, different from RPE in convolution, DRPE in attention mechanism considers the relationship between content and position feature, which is demonstrated effective in previous works~\cite{Yang2019XLNetGA,He2020DeBERTaDB}. Our experiments indicated that the relative position information is vital for sequence-to-sequence mapping learning. 

\begin{figure*}[t]
\centering
\includegraphics[width=0.9\textwidth,height=0.33\textheight]{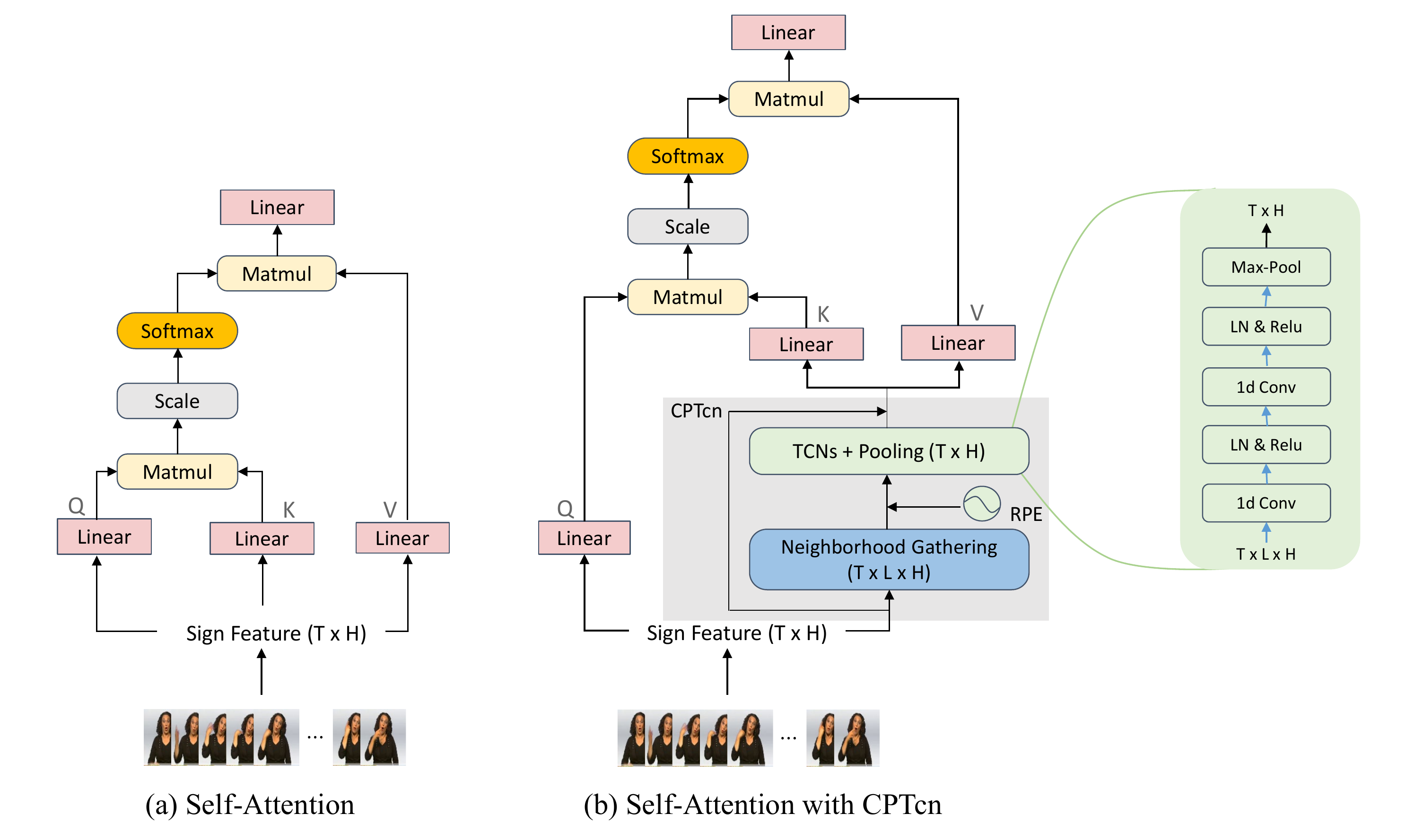}
\caption{Vanilla self-attention and self-attention equipped with Content-aware and Position-aware Temporal Convolution (CPTcn). Where neighborhood gathering denotes selecting adjacent relevant features in a contiguous local region, TCNs denote stacked temporal convolution layers.}
\label{convattention} 
\end{figure*} 

\section{Method}\label{Method}

\subsection{Preliminaries and Model Overview}
Figure~\ref{architecture} illustrates the overall architecture of our proposed model, which jointly learns to recognize and translate sign videos into gloss annotations and spoken language sentences. In the following subsections, we will first revisit the sign language Transformer structure and then give detailed descriptions about our proposed two methods: content-aware and position-aware temporal convolution (CPTcn), and self-attention with disentangled relative position encoding (DRPE). 

\subsection{Joint Sign Language Transformer Structure}

Given a series of video frames, the vanilla sign language Transformer (SLTR) model firstly adopts a CNN backbone to extract frame-wise spatial features and uses a word embedding to transfer one-hot natural language words into dense vectors. Then a Transformer-based encoder-decoder model is utilized to learn SLR and SLT simultaneously. For SLR, the encoder output learned temporal sign features. A Connectionist Temporal Classification (CTC)~\cite{Graves2006ConnectionistTC} loss is applied to learn the mapping of gloss annotations and sign features. For SLT, the decoder output decomposes sequence level conditional probabilities in an autoregressive manner and then calculates the cross-entropy loss for each word. Meanwhile, the learning of SLR and SLT share the Transformer encoder.  

Vanilla Transformer is a sequence-to-sequence structure, which consists of several Transformer blocks. Each block contains a multi-head self-attention and a fully feed-forward network. Given a feature sequence \(F \in R^{M\times d}\) with $M$ frames, taking single-head attention as an example, the standard self-attention can be formulated as:

\begin{equation}
\begin{aligned}
& Q=FW_q, K=FW_k, V=FW_v, \\
& S = QK^T, \\
& \text{Attn}(Q,K,V) = \textup{softmax}(\dfrac{S}{\sqrt{d}})V, \\
& a_{ij} = \dfrac{exp(s_{ij}/\sqrt{d})}{\sum_{j'}exp(s_{ij'}/\sqrt{d})} \\
\end{aligned}
\label{eqn:equation1}
\end{equation}

\noindent where \(W_q,W_k,W_v \in R^{d\times d}\) represents projection matrices.   $S\in R^{M\times M}$ represents the similarity computed by query $Q\in R^{M\times d}$ and key $K\in R^{M\times d}$. $a_{ij}$ represent the normalized attention weights respectively.

Our work concentrates on improving the self-attention mechanism to understand sign video and target sentences better. To focus on our main contributions, we omit the detailed architecture and refer readers to ~\cite{Vaswani2017AttentionIA} for reference.

\subsection{Content-aware and Position-aware Temporal Convolution}
As shown in Figure~\ref{convattention}, we propose content-aware and position-aware temporal convolution (CPTcn) to learn local temporal semantics, aiming at obtaining more discriminative sign representations. In this section, we first introduce a content-aware neighborhood gathering method, which adaptively selects surrounding neighbors. Secondly, we elaborate on the detail of endowing the temporal convolution with relative position information, which models the relationship between surrounding features and the current feature. Finally, we incorporate the proposed CPTcn module with the self-attention mechanism.

\begin{figure*}[t]
\centering
\includegraphics[width=0.96\textwidth,height=0.27\textheight]{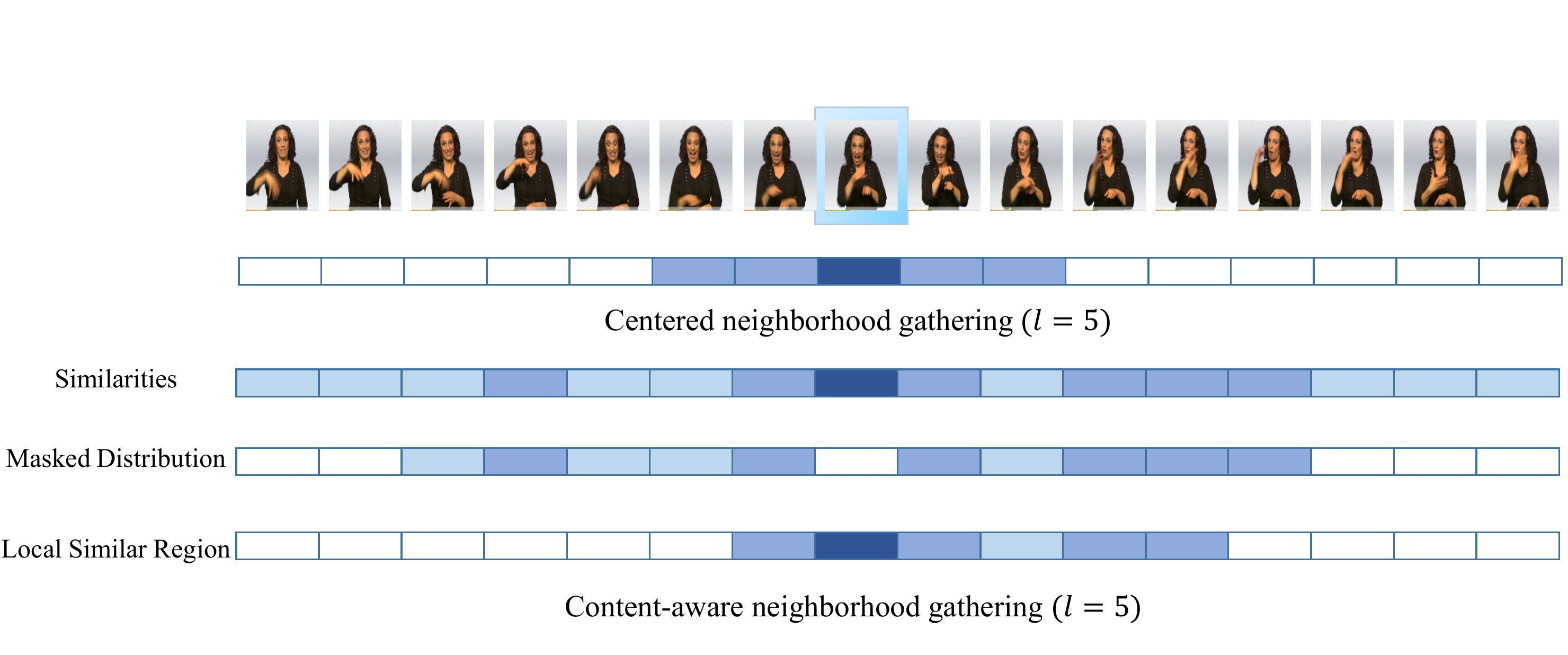}
\caption{Our proposed content-aware neighborhood gathering method, compared with the centered neighborhood gathering method.}
\label{localregionselect} 
\end{figure*}

\subsubsection{Content-aware neighborhood gathering Method} In sign videos, we observe that each sign gesture usually lasts about 0.5\(\sim\)0.6 seconds (\(\sim\)16 frames). However, the vanilla Sign Language Transformer (SLTR) model aggregates sign features in a frame-wise manner, thus neglecting the local temporal structure of sign gestures. Unlike their work, we develop a content-aware neighborhood gathering method to adaptively select the relevant surrounding features, which are around a specific feature and in a contiguous region. Shown as Figure~\ref{localregionselect}, we obtain the clip-level feature with neighboring features via three steps:

\noindent \textbf{1).} Given the sequential representations $F=\{f_1,f_2, ...,f_M\}$ from the CNN backbone model, we apply outer tensor product to get a similarity matrix $s \in R^{M\times M}$:

\begin{equation}
\begin{aligned}
s = \frac{FF^{T}}{\sqrt{d}}
\end{aligned}
\label{eqn:equation2}
\end{equation}

\noindent where the diagonal elements in $s$ represent similarities towards the features themselves.

\noindent \textbf{2).} To ensure neighbors are going to be selected instead of the far-away ones, we only consider a range $[t-l, t+l]$ for a specific feature $f_t\in{R^{d}}$ to keep local semantic consistency. Then we replace the similarity scores with -\textit{inf} outside this range and at the current feature. Mathematically, the selecting criterion for $f_t$ becomes as:

\begin{equation}
\begin{aligned}
s_{t,j} = \begin{cases} 
s_{t,j}, & \text{$j \in [\max(0, t-l), t)$ \& $(t, \min(t+l, M)]$} \\
-inf, & \text{others}
\end{cases}
\end{aligned}
\label{eqn:equation3}
\end{equation}

\noindent where $l$ represents the maximum distance among the considering features from the current feature. Then we apply the softmax function to obtain the masked distribution $d^{t} \in R^{M}$ in the local region around the current feature $f_t$:

\begin{equation}
d_{t} = \text{softmax}(s_{t})
\label{eqn:equation4}
\end{equation}

\noindent Note that the weight at the current feature is zero, thus the summation of the weights before and after the current feature is 1.

\noindent \textbf{3).} It is hard to determine the size and boundaries of the local region. Fortunately, the normalized distribution of similarities obtained in Equation~\ref{eqn:equation4} indicates the location of similar neighbors. Therefore, we use the weights of the normalized distribution before and after the current feature to adaptively determine the size of the selected region. Respectively, we define the size before and after the current feature with $l_{-}$ and $l_{+}$:

\begin{equation}
\begin{aligned}
l_{-} = \gamma \sum d_{t,j}\cdot l&, \quad \text{$\max(0, t-l)\leq j < t$} \\
l_{+} = \gamma \sum d_{t,j}\cdot l&, \quad \text{$t < j \leq \min(t+l, M)$} \\
l_r = l_{-} &+ l_{+} = \gamma \cdot l \\
\end{aligned}
\label{eqn:equation5}
\end{equation}

\noindent where $\gamma$ is a hyperparameter to control the size of selected region, and the size of the region is $l_r$. We define the final selected contiguous region as $LSR_t$ (\textbf{L}ocally \textbf{S}imilar \textbf{R}egion) for a specific feature ${f_t}$:

\begin{equation}
\begin{aligned}
\text{LSR}_t = \{f_{t-l_{-}}, .., f_t, ..., f_{t+l_{+}}\}
\end{aligned}
\label{eqn:equation6}
\end{equation}

\noindent Finally, we adaptively obtain the clip-level features which are in a contiguous local region:

\begin{equation}
\begin{aligned}
f_{t}^{r} = \mathcal{F}_{CNG}(f_t, LSR_t)
\end{aligned}
\label{eqn:equation7}
\end{equation}
\noindent where $\mathcal{F}_{CNG}$ denotes the content-aware neighborhood gathering method, and $f_t^r \in R^{l_r\times d}$ denotes the current feature $f_t$ with its $l_r$ surrounding neighbors. The clip-level features with temporal surrounding neighbors can be computed using Algorithm~\ref{alg:algorithm1}.

\begin{algorithm}[htb]
  \caption{Content-aware neighborhood gathering Method.}
  \label{alg:algorithm1}
  \begin{algorithmic}[1]
    \Require
      Frame-wise spatial feature $F\in R^{M\times d}$ from CNN backbone;
    \Ensure Clip-level features $F^r \in R^{M\times l_r \times d}$;
    \State $s = \frac{1}{\sqrt{d}}ff^{T}$;
    \label{code:fram:similarity}     
    \State $s=\text{local_mask}(s, l-1, -\text{inf})$;
    \label{code:fram:localmask}
    \State $s=\text{diagonal_mask}(s, -\text{inf})$;
    \label{code:fram:diagonalmask}
    \State $a = \text{softmax}(s, \text{dim}=-1)$;
    \label{code:fram:softmax}
    \State{$F^r = []$}
    \For {$t=0,...,M-1$}
        \State {$l_{-}=sum(a[max(0, t-l):t]) * l * \gamma$} 
        \State {$l_{+}=sum(a[t, t+1:min(t+l, M)]) * l * \gamma$}
        \State {$\text{inds} = [t-l_{-}, .., t, .., t+l_{+}]$}
        \State{$\text{neighbors} = \text{index_select}(f_{t}, \text{inds})$}
        \State{$F^r \text{ append neighbors}$}
    \EndFor
    \State{$\text{concatenate}(F^r, \text{dim}=0)$}  \\
    \Return $F^r$; 
  \end{algorithmic}
\end{algorithm}

\subsubsection{Position-aware Temporal Convolution}\label{tcrpe}
Temporal convolution is a common method to aggregate sequential features. However, convolution layers with a limited receptive field are insufficient to capture the position information~\cite{Islam2020HowMP}, which is important for sign gesture understanding. More specifically, the recognition of sign language is sensitive to the frame order. Absolute position encoding used in previous methods~\cite{Camgz2020SignLT,Li2020TSPNetHF} is a promising approach to encode position information. However, it is demonstrated direction- and distance-unaware~\cite{Yan2019TENERAT}. Inspired by recent work on language modelling~\cite{Raffel2020ExploringTL}, we infuse relative position information to the clip-level feature. We first compute the relative position matrix $R\in R^{M\times l_r}$ between the frame-wise feature and the current feature:

\begin{equation}
\begin{aligned}
R_{t,p} = p-t,\quad \text{$t\in [0, M-1]$, $p\in (t-l_{-}, t+l_{+}]$}.
\end{aligned}
\label{eqn:equation8}
\end{equation}

Then we represent the relative position indices in learnable embedding space, and obtain the position embeddings $\Phi_{rpe}\in R^{M\times l_r \times d}$. Adding $\Phi_{rpe}$ to clip-level features $F^r$, resulting in position-informed clip-level representation:

\begin{equation}
\begin{aligned}
\hat{F^r} = F^r + \Phi_{rpe}.
\end{aligned}
\label{eqn:equation9}
\end{equation}

Lastly, we aggregate the clip-level features with position information to compressed features $F_{ag}\in R^{M\times d}$, and apply a residual function:

\begin{equation}
\begin{aligned}
F_{ag}=\text{MaxPool}(\text{Relu}(\text{LN}(\text{Conv1d}(\hat{F^r})))) + F
\end{aligned}
\label{eqn:equation10}
\end{equation}

\noindent where $\text{LN}$ represents the Layer Normalization~\cite{Ba2016LayerN}. Conv1d(·) performs a 1-D convolution filter on time dimension with Relu(·) activation function. As shown in Figure~\ref{convattention}, we use two layers of such a network, which is omitted here for readability.

\subsubsection{Self-attention with CPTcn}
Similar to the vanilla Transformer model, we feed the aggregated feature $F_{ag}$ to the self-attention mechanism. Note that, as shown in Figure~\ref{convattention}, we only set $K=V=F_{ag}$, and keep $Q$ as original frame-wise representation $F$. The reason for this design is to maintain the difference within adjacent features in $Q$. Experimental result demonstrates that this network design performs better than $Q=K=V=F_{ag}$.

\subsection{Self-Attention with DRPE}
As shown in Figure~\ref{architecture}, we further inject relative position information into the attention mechanism for sign video learning, target sentence learning, and mapping learning between them. Most existing approaches for endowing the attention mechanism relative position information are based on pairwise distance~\cite{Shaw2018SelfAttentionWR}. They have been explored in machine translation ~\cite{Shaw2018SelfAttentionWR}, music generation~\cite{Huang2019MusicTG} and language modelling~\cite{Dai2019TransformerXLAL,He2020DeBERTaDB}. Here, we propose a  disentangled relative position encoding (DRPE)~\cite{He2020DeBERTaDB}. 

Different from RPE used in Section~\ref{tcrpe}, DRPE considers the correlations between relative positions and content features, which are proven that improving the performance~\cite{He2020DeBERTaDB, Yang2019XLNetGA}. Specifically, we separate the content features and relative position encoding to compute attention weights. The first line of projection in Equation~\ref{eqn:equation1} is reparameterized as:

\begin{equation}
\begin{aligned}
& Q_f=FW_{q,c},K_f=FW_{k,c},V_f=FW_{v,c} \\
& Q_p=PW_{q,p},K_p=PW_{k,p}
\end{aligned}
\label{eqn:equation11}
\end{equation}

\noindent where $F\in R^{M\times d}$ represent the content feature. $Q_f, K_f, V_f\in R^{M\times d}$ represent query, key and value content vectors which are obtained with projection matrices $W_{q,c},W_{k,c},W_{v,c}\in R^{d\times d}$. $P\in R^{2 \mathcal{L} \times d}$ represents created learned relative position embedding, where $\mathcal{L}$ is the max relative distance. $Q_p, K_p \in R^{ 2 \mathcal{L} \times d}$ represent the projected position embedding with projection matrices $W_{q,p}, W_{k,p}\in R^{d\times d}$, respectively.

Following this, we generate the attention weights with the relative position bias. The calculation of pairwise \textit{content-content} is in the same way as standard self-attention, thereby generating the content-based content vector. While the calculation of pairwise \textit{content-position} is different from standard self-attention. We first create a relative position distance matrix $R^{rel}\in R^{M\times M}$, and then generate the position-based content vectors. The $2\sim 4$ lines of computing attention weights in Equation~\ref{eqn:equation1} are reparameterized as:

\begin{equation}
\begin{aligned}
s_{i,j}^{rel} =& \{Q_{f,i}, \textcolor{blue}{Q_pR^{rel}_{i-j}}\} \times \{K_{f,j}, \textcolor{blue}{K_pR^{rel}_{j-i}}\}^T \\
=& \begin{matrix} \underbrace{ Q_{f,i}K_{f,j}^T } \\ \textit{c2c} \end{matrix} + 
   \begin{matrix} \underbrace{ Q_{f,i}\textcolor{blue}{{K_p}^T{R^{rel}_{j-i}}^T} } \\ \textit{c2p} \end{matrix} + 
   \begin{matrix} \underbrace{\textcolor{blue}{Q_pR^{rel}_{i-j}}K_{f,j}^T } \\ \textit{p2c} \end{matrix} + \\
   \;& \begin{matrix} \underbrace{ \textcolor{blue}{Q_pR^{rel}_{i-j}{K_p}^T{R^{rel}_{j-i}}^T} } \\ \textit{p2p} \end{matrix} \\
\textup{Attn}(&Q_f,K_f,V_f, \textcolor{blue}{Q_pR^{rel}_{q-k}}, \textcolor{blue}{K_pR^{rel}_{k-q}})=\textup{softmax}(\dfrac{S^{rel}}{\sqrt{4d}})V_f,\\
a_{ij} =& \dfrac{exp(s^{rel}_{ij}/\sqrt{4d})}{\sum_{j'}exp(s^{rel}_{ij'}/\sqrt{4d})} \\
\end{aligned}
\label{eqn:equation12}
\end{equation}

\noindent where $S^{rel}\in R^{M \times M}$ represents the unnormalized attention score matrix and $s^{rel}_{ij}$ represents the score computed by query at position $i$ and key at the position $j$. $R^{rel}_{q-k}\in R^{M\times M}$ represents the relative distance matrix computed by the positions of query and key. $R^{rel}_{i-j}$ lies in the $(i,j)$-th of $R^{rel}_{q-k}$, and represents the relative distance between $i$-th query and $j$-th key. $R^{rel}_{k-q}\in R^{M\times M}$ and $R^{rel}_{j-i}$ are computed in similar ways. Note that $R^{rel}_{i-j}$ and $R^{rel}_{j-i}$ are opposite numbers thus providing our model with directional information.

Moreover, in the first line of the above equation, the first item $c2c$ represents $\textit{content-to-content}$ which is the content-based content vectors. The second and third item $c2p$ and $p2c$ represent $\textit{content-to-position}$ and $\textit{position-to-content}$ respectively, which are relative position based content vectors. $p2p$ represents $\textit{position-to-position}$ which is omitted in vanilla DRPE~\cite{He2020DeBERTaDB}. However, in our experiments, we find that $p2p$ bring improvements to our performance in both recognition and translation. Therefore, we keep this item of \textit{position-to-position}. In Section~\ref{analysisdrpe}, we analyze the impact of different item in the first line of Equation~\ref{eqn:equation12}.

Preceding this, in the last two lines, we apply softmax function and scaling factor $\frac{1}{\sqrt{4d}}$ to get normalized scaled attention weights. 

Totally, there are two differences between the DRPE method applied in our architecture and DeBERTa~\cite{He2020DeBERTaDB}. The first is that we consider the position-to-position information, which is omitted in DeBERTa. Experimental results in Table~\ref{itemindrpe} show the effectiveness of this item. The second difference is that DRPE is used in text-only in DeBERTa for language modeling. However, in our proposed model, as seen in Figure~\ref{architecture}, we apply the relative position method in text-only target sentence learning, image-only sign video learning, and even the cross-modal video sequence and target sentence interaction. Experimental results in Table~\ref{drpeinsa} show the effectiveness of our improvements. Note that we are the first to consider the relative position relationship between sign frames and target words.

In summary, equipping with the CPTcn module and DRPE in self-attention layers, the heart module in the Transformer model, we finally arrive at our proposed PiSLTRc model.

\section{Experiments}\label{Experiments}
\subsection{Dataset and Metrics}
We evaluate our method on three datasets, including PHOENIX-2014~\cite{Forster2014ExtensionsOT}, PHOENIX-2014-\textbf{T}~\cite{Camgz2018NeuralSL} and Chinese Sign Language (CSL)\cite{Huang2018VideobasedSL}.

\textbf{PHOENIX-2014} is a publicly available German Sign Language dataset, which is the most popular benchmark for continuous SLR. The corpus was recorded from broadcast news about weather. It contains videos of 9 different signers with a vocabulary size of 1295. The split of videos for Train, Dev, and Test is 5672, 540, and 629, respectively.

\textbf{PHOENIX-2014-T} is the benchmark dataset of sign language recognition and translation. It is an extension of the PHOENIX14 dataset~\cite{Forster2014ExtensionsOT}. Parallel sign language videos, gloss annotations, and spoken language translations are available in PHOENIX14\textbf{T}, which makes it feasible to learn SLR and SLT tasks jointly. The corpus is curated from a public television broadcast in Germany, where all signers wear dark clothes and perform sign language in front of a clean background. Specifically, the corpus contains 7096 training samples (with 1066 different sign glosses in gloss annotations and 2887 words in German spoken language translations), 519 validation samples, and 642 test samples. 

\textbf{CSL} is a Chinese Sign Language dataset, which is also a widely used benchmark for continuous SLR. These videos were recorded in a laboratory environment, using a Microsoft Kinect camera with a resolution of 1280 × 720 and a frame rate of 30 FPS. In this corpus, there are 100 sentences, and each sentence is signed five times by 50 signers (in total 2,500 videos). As no official split is provided, we split the dataset by ourselves. We give 20,000 and 5,000 samples to the training set and testing set, respectively. When splitting the dataset, we ensure that the sentences in the training and testing sets are the same, but the signers are different.

We evaluate our model on the performance of SLR and SLT as following~\cite{Camgz2018NeuralSL}:

\textbf{\textit{Sign2gloss}} aims to transcribe sign language videos to sign glosses. It is evaluated using word error rate (WER), which is a widely used metric for cSLR:
\begin{equation}
\begin{aligned}
\text{WER} = \dfrac{\#\text{substitution} + \#\text{deletion} + \#\text{insertion}}{\#\text{words in reference}}
\end{aligned}
\label{eqn:equation13}
\end{equation}

\textbf{\textit{Sign2text}} aims to directly translate sign language videos to spoken language translation without intermediary representation. It is evaluated using BLEU~\cite{Papineni2002BleuAM} which is widely used for machine translation.

\textbf{\textit{Sign2(gloss+text)}} aims to jointly learn continuous SLR and SLT simultaneously. This approach is currently state-of-the-art in the performance of SLT since the training of cSLR brings benefits for sign video understanding, thus improving the performance of translation.

\subsection{Implementation and Evaluation Details}

\subsubsection{Network Details} Like Camgz \textit{et al.}~\cite{Camgz2020SignLT}, we extract frame-wise spatial sign features with CNN backbone from CNN-LSTM-HMM~\cite{Koller2020WeaklySL}. Then we apply the improved Transformer network to learn SLR and SLT simultaneously. Its setting used in our experiments is based on Camgz \textit{et al.}~\cite{Camgz2020SignLT}. Specifically, we use 512 hidden units, 8 heads, 6 layers, and 0.1 dropout rate.

In our proposed CPTcn model, the size of the select contiguous local similar region $l_r$ is set to be 16 (about 0.5-0.6 seconds), which is the average time needed for completing a gloss. We analyze the impact of the size in Section~\ref{analysisofng}.

The setting of two temporal convolution layers is F3-S1-P0-F3-S1-P0, where F, S, P denote the kernel filter size, stride, and padding size, respectively. The analysis of different modules of the position-informed convolution is concluded in Section~\ref{analysisofptcn}.

In the self-attention and cross-attention mechanism, we apply DRPE to inject relative position information. We set the max relative distance $\mathcal{L}$ to be 32 in our experiments. The analysis of the DRPE is conduct in Section~\ref{analysisdrpe}.

Besides, we train the SLR and SLT simultaneously. Thus we set $\lambda_R$ and $\lambda_T$ as the weight of recognition loss and translation loss. 

\subsubsection{Training} We use the Adam optimizer~\cite{Kingma2015AdamAM} to optimize our model. We adopt the warmup schedule for learning rate that increases the learning rate from 0 to 6.8e-4 within the first 4000 warmup steps and gradually decay it with respect to the inverse square root of training steps. We train the model on 1 NVIDIA TITAN RTX GPU, and use 5 checkpoints averaging for the final results.

\subsubsection{Decoding} During inference, we adopt CTC beam search decoder with a beam size of 5 for SLR decoding. Meanwhile, we also utilize the beam search with the width of 5 for SLT decoding, and we apply a length penalty~\cite{Wu2016GooglesNM} with $\alpha$ values ranging from 0 to 2.

\subsection{Ablation Study}

\begin{table}[t]
\renewcommand\arraystretch{1.2}
\centering
\smallskip
\caption{Evaluation of different neighborhood gathering method on PHOENIX-2014-T. "NG" is the abbreviation of neighborhood gathering.}
\resizebox{0.48\textwidth}{!}{
\begin{tabular}{c|cc|cc}
\hline
\multirow{2}{*}{NG method} & \multicolumn{2}{c|}{SLR(WER)} & \multicolumn{2}{c}{SLT(BLEU-4)} \\ \cline{2-5}
 & DEV & TEST & DEV & TEST \\
\hline
w/o NG & 24.23 & 24.92& 20.54  &  20.80  \\
Centered NG ($l_r=16$) & 23.64 & 24.17 & 20.73  &  21.23  \\
Sparse NG ($l_r=16$) & 23.06 & 23.52 & 21.83 &   22.08 \\
Content-aware NG ($l_r=16$) & \textbf{22.23} & \textbf{23.01} & \textbf{23.17}  & \textbf{23.40} \\
\hline
\end{tabular}}
\label{shapeofasm}
\end{table}

\begin{table}[t]
\renewcommand\arraystretch{1.2}
\centering
\smallskip
\caption{Evaluation of the size of the selected local similar region on PHOENIX-2014-T.}
\resizebox{0.4\textwidth}{!}{
\begin{tabular}{c|cc|cc}
\hline
\multirow{2}{*}{Size of LSR} & \multicolumn{2}{c|}{SLR(WER)} & \multicolumn{2}{c}{SLT(BLEU-4)} \\ \cline{2-5}
 & DEV & TEST & DEV & TEST \\
\hline
$l_r=8$ & 22.85 & 23.85 & 22.30  & 22.99 \\
$l_r=12$ & 22.52 & 23.62 & 22.31  & \textbf{23.51}  \\
$l_r=16$ & \textbf{22.23} & \textbf{23.01} & \textbf{23.17}  & 23.40  \\
$l_r=20$ & 23.02 & 23.74 & 22.85  & 22.86  \\
\hline
\end{tabular}}
\label{sizeoflsr}
\end{table}

\subsubsection{Analysis of content-aware neighborhood gathering method}\label{analysisofng} In our proposed CPTcn module, we introduce a content-aware neighborhood gathering method to select the relevant surrounding neighbors dynamically. Three potential concerns with using this method are: 1) How many improvements does the content-aware method bring? 2) Must be the selected features contiguous in position? 3) What is the appropriate size of the selected region?

In Table~\ref{shapeofasm}, we compared three methods to verify the first two questions: the essential of whether the selected region is content-aware and contiguous. For notation, \textbf{w/o NG} means no neighborhood gathering method. \textbf{Centered NG} means directly to select k features centered around the current feature. \textbf{Sparse NG} means dynamically selecting k features with the highest similarity, which may be discontinuous in position.  \textbf{Content-aware NG} means to select k contiguous features adaptively based on similarity using our proposed content-aware segmentation method. We can see those neighborhood gathering methods effectively improve the performance. \textbf{Sparse NG} substantially outperforms \textbf{Centered NG}. This gap suggests that the content-aware method is critical for feature selecting. Moreover, \textbf{Content-aware NG} performs better than other methods. This indicates that our content-aware contiguous feature aggregation is more suitable for capturing sign gesture representation. 

In Table~\ref{sizeoflsr}, we explore the appropriate size of the select local similar region (LSR). The performance of our model performs best when the size of LSR is 16. This is consistent with the finding that the 16-frame (about  0.5-0.6  seconds) is the average time needed for completing a gloss. Besides, by gathering the larger width regions (for example, 20 frames), we observed slight performance degradation. This is because 20 frames (about 1 second) usually contain more than one gesture and thus lower the performance.

\begin{table}[t]
\renewcommand\arraystretch{1.2}
\centering
\smallskip
\caption{Evaluation of different model in CPTcn module on PHOENIX-2014-T. "PE" denotes position encoding. "APE" denotes absolute position encoding.}
\resizebox{0.4\textwidth}{!}{
\begin{tabular}{c|cc|cc}
\hline
\multirow{2}{*}{module in CPTcn} & \multicolumn{2}{c|}{SLR (WER)} & \multicolumn{2}{c}{SLT (BLEU-4)} \\ \cline{2-5}
 & DEV & TEST & DEV & TEST \\
\hline
w/o PE & 23.44 & 24.03 & 21.15 & 21.47 \\
w/ APE & 22.89 & 23.47 & 21.89 & 22.18   \\
w/ RPE & \textbf{22.23} & \textbf{23.01} & \textbf{23.17}  & \textbf{23.40} \\
\hline \hline
w/o Redisual & 24.19 & 25.31 & 20.78 & 20.36 \\
w/o LN & 23.01 & 23.72 & 21.99 & 21.57 \\
\hline
CPTcn & \textbf{22.23} & \textbf{23.01} & \textbf{23.17}  & \textbf{23.40} \\
\hline
\end{tabular}}
\label{peoftcn}
\end{table}

\subsubsection{Analysis of position-aware Temporal Convolution}\label{analysisofptcn} In the first two lines in Table~\ref{peoftcn}, we study the relative position encoding in the CPTcn module. Experimental results show that position information is crucial for aggregating the sequential features. Furthermore, compared with absolute position encoding (APE), relative position encoding (RPE) bring improvements with $+1.22$ BLEU scores and $-0.46\%$ WER score on the test dataset. The result supports the conjecture of Yan \textit{et. al.}~\cite{Yan2019TENERAT} that RPE provides direction and distance awareness for sequence modeling compared with APE method. 

Moreover, we explore the network design of the temporal feature aggregator method in the $3\sim 4$ lines in Table~\ref{peoftcn}. Experiments show that residual connection is essential. And Layer normalization is effective for sequential feature modeling.

\begin{figure*}[t]
\centering
\includegraphics[width=1.0\textwidth]{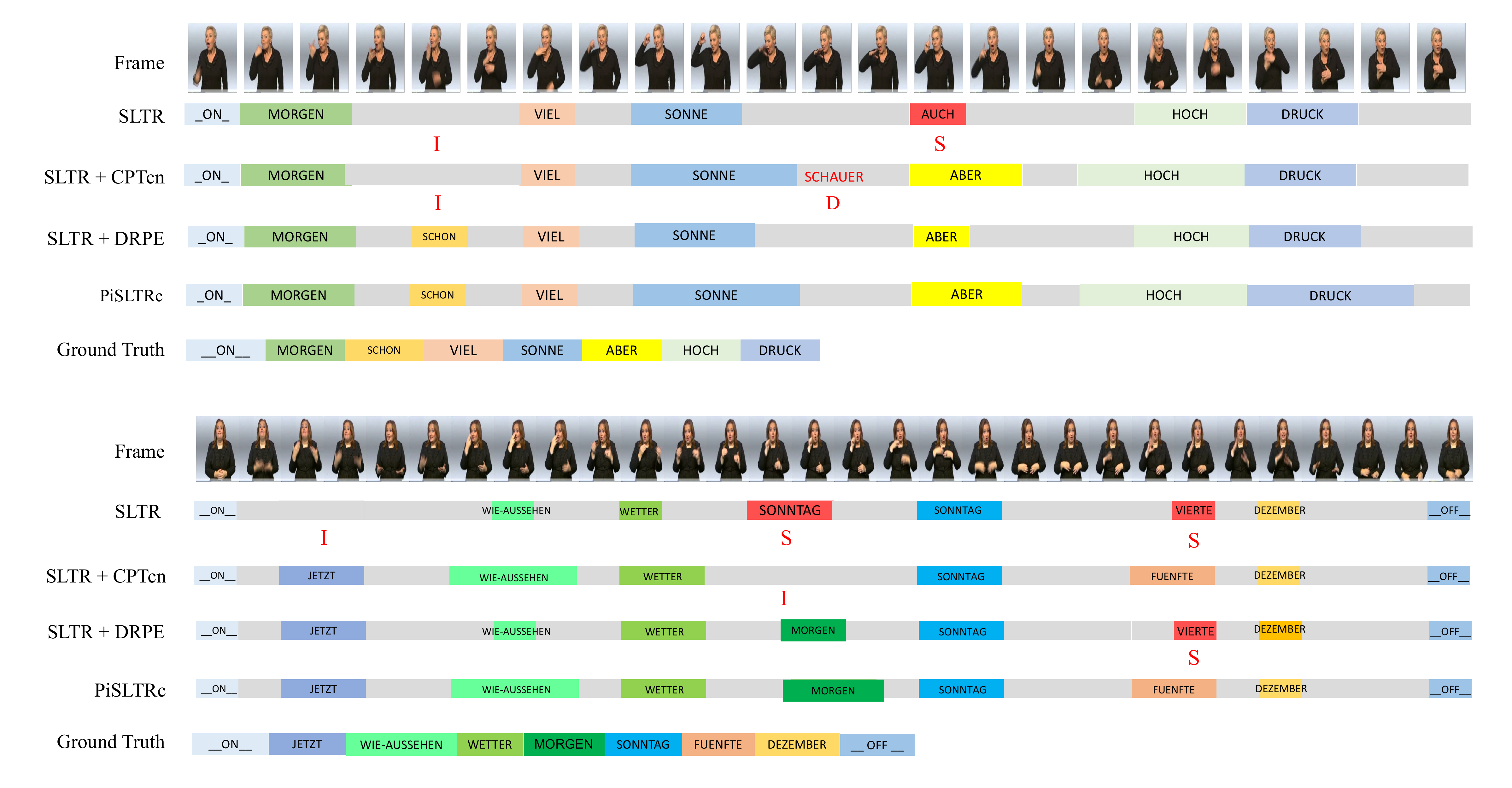}
\caption{Qualitative recognition results of our proposed modules from Dev set (D: delete, I: insert, S: substitute).}
\label{testexample} 
\end{figure*}

\subsubsection{Analysis of DRPE in self-attention}\label{analysisdrpe} We further conduct comparative experiments to analyze the effectiveness of disentangle relative position encoding (DRPE) in the attention mechanism. As shown in Figure~\ref{architecture}, we replace absolute position encoding (APE) with DRPE in three places: encoder self-attention, decoder self-attention, and encoder-decoder cross attention. For notation, in Table~\ref{drpeinsa}, "Enc-SA" means self-attention in the encoder module. "Dec-SA" means self-attention in the decoder module. "Enc-Dec-CA" means cross attention between encoder and decoder. In the $1\sim 4$ lines of Table~\ref{drpeinsa}, we can see that the DRPE method used in the encoder and decoder all brings significant improvements. This further demonstrates that relative position encoding provides the direction and distance awareness for sequence representation learning. In addition, we find that the performance of DRPE used only in the encoder is better than that of DRPE used only in the decoder. This phenomenon suggests that direction and distance information are more critical for sign video learning than sentence representation learning.

\begin{table}[t]
\renewcommand\arraystretch{1.2}
\centering
\smallskip
\caption{Analysis of DRPE in encoder self-attention and decoder self-attention.”SA” is the abbreviation of self-attention. "CA" means the cross-attention of encoder-decoder.}
\resizebox{0.48\textwidth}{!}{
\begin{tabular}{c|cc|cc}
\hline
\multirow{2}{*}{Method} & \multicolumn{2}{c|}{SLR (WER)} & \multicolumn{2}{c}{SLT (BLEU-4)} \\ \cline{2-5}
 & DEV & TEST & DEV & TEST \\
\hline
 w/ APE & 25.36 & 25.27 & 20.12  &  20.39   \\
Enc-SA w/ DRPE & 22.89 & 23.76 & 22.35 & 22.47  \\
Dec-SA w/ DRPE & 23.29 & 23.84 & 21.89 & 21.27 \\
Enc-SA \& Dec-SA w/ DRPE & 22.54 & 22.89 & 22.78 & 22.90  \\
\hline
Enc-Dec-CA w/ DRPE & 23.74 & 23.93 & 21.59 & 21.41 \\
\hline
All w/ DRPE & \textbf{22.23} & \textbf{23.01} & \textbf{23.17}  & \textbf{23.40} \\
\hline
\end{tabular}}
\label{drpeinsa}
\end{table}

\begin{table}[t]
\renewcommand\arraystretch{1.2}
\centering
\smallskip
\caption{Analysis of different item in DRPE. "c2c" denotes \textit{content-to-content}, "c2p" denotes \textit{content-to-position}, "p2c" denotes \textit{position-to-content}, and "p2p" denotes \textit{position-to-position}.}
\resizebox{0.42\textwidth}{!}{
\begin{tabular}{l|cc|cc}
\hline
\multirow{2}{*}{Item in DRPE} & \multicolumn{2}{c|}{SLR (WER)} & \multicolumn{2}{c}{SLT (BLEU-4)} \\ \cline{2-5}
 & DEV & TEST & DEV & TEST \\
\hline
c2c only & 25.79 & 25.85 & 20.03  &  20.18   \\
$\quad$ + c2p \& p2c & 22.57 & 23.26 & 22.84 & 22.79  \\
$\quad$ + p2p & \textbf{22.23} & \textbf{23.01} & \textbf{23.17}  & \textbf{23.40} \\
\hline
\end{tabular}}
\label{itemindrpe}
\end{table}

As we move to the fourth line in Table~\ref{drpeinsa}, the results show that DRPE in encoder-decoder attention also increases the performance. This phenomenon shows that even if the order of the word in the natural language is inconsistent with the sign language gloss, the relative position information still benefits their mapping learning.

Different from DRPE used in DeBERTa ~\cite{He2020DeBERTaDB}, we further explore the effectiveness of different items mentioned in Equation~\ref{eqn:equation12} in our task. Experimental results are shown in Table~\ref{itemindrpe}, the correlations between content and position feature bring significant improvement. Moreover, the \textit{position-to-position} item also benefits our model. This result is consistent with the conclusion in Ke \textit{et al.}~\cite{Ke2020RethinkingPE}. Accordingly, we adopt these four items in our disentangled relative position encoding.

\begin{table*}[t]
\renewcommand\arraystretch{1.1}
\centering
\smallskip
\caption{Qualitative results with different methods on SLT task.}
\resizebox{1.0\textwidth}{0.18\textheight}{
\begin{tabular}{r|l}
\hline
GT: & in der nacht sinken die temperaturen auf vierzehn bis sieben grad . \\
&  (at night the temperatures drop to fourteen to seven degrees .) \\
SLTR: & heute nacht werte zwischen sieben und sieben grad . \\
& (tonight values between seven and seven degrees .) \\
PiSLTRc: & heute nacht kühlt es ab auf vierzehn bis sieben grad . \\
& (tonight it's cooling down to fourteen to seven degrees .)\\
\hline
GT: & an der saar heute nacht milde sechzehn an der elbe teilweise nur acht grad . \\
 & (on the saar tonight a mild sixteen on the elbe sometimes only eight degrees .) \\
SLTR: & südlich der donau morgen nur zwölf am oberrhein bis zu acht grad . \\
& (south of the danube tomorrow only twelve on the upper rhine up to eight degrees .) \\
PiSLTRc: & am oberrhein heute nacht bis zwölf am niederrhein nur kühle acht grad . \\
& (on the upper rhine tonight until twelve on the lower rhine only a cool eight degrees .) \\
\hline
GT: & am tag von schleswig holstein bis nach vorpommern und zunächst auch in brandenburg gebietsweise länger andauernder regen . \\
 & (In the south, denser clouds sometimes appear, otherwise it is partly clear or only slightly cloudy .) \\
SLTR: & am mittwoch in schleswig holstein nicht viel regen . \\
 & (not much rain on wednesday in schleswig holstein .) \\
PiSLTRc: & am donnerstag erreicht uns dann morgen den ganzen tag über brandenburg bis zum teil dauerregen .\\
 & (on thursday we will reach us tomorrow the whole day over brandenburg until partly constant rain.) \\
\hline
GT: & im süden gibt es zu beginn der nacht noch wolken die hier und da auch noch ein paar tropfen fallen lassen sonst ist es meist klar oder nur locker bewölkt . \\
& (In the south there are still clouds at the beginning of the night that drop a few drops here and there, otherwise it is mostly clear or only slightly cloudy .) \\
SLTR: & im süden tauchen im süden teilweise dichtere wolken auf sonst ist es verbreitet klar . \\
& (in the south there are sometimes denser clouds in the south otherwise it is widely clear .) \\
PiSLTRc: & im süden tauchen auch mal dichtere wolken auf sonst ist es gebietsweise klar oder nur locker bewölkt . \\
& (In the south, denser clouds sometimes appear, otherwise it is partly clear or only slightly cloudy .) \\
\hline
\end{tabular}}
\label{exampleslt}
\end{table*}

\begin{table}[t]
\renewcommand\arraystretch{1.3}
\centering
\smallskip
\caption{The evaluation results on sign2gloss task on PHOENIX-2014-T dataset.}
\resizebox{0.48\textwidth}{!}{
\begin{tabular}{c|cc|cc}
\hline
\textit{sign2gloss} & \multicolumn{2}{c|}{DEV} & \multicolumn{2}{c}{TEST} \\
\hline
Model & del/ins & WER & del/ins & WER \\
\hline
DNF~\cite{Cui2019ADN} & 5.9/3.0 & 22.7 & 6.8/2.9 & 23.4 \\
CNN+LSTM+HMM~\cite{Koller2020WeaklySL} & - & 24.5 & - & 26.5  \\
SLTR-R~\cite{Camgz2020SignLT} & - & 24.9 & - & 24.6 \\
FCN~\cite{Cheng2020FullyCN} & 6.5/3.2 & 22.9 & 5.8/4.7 & 23.7 \\
STMC (RGB) ~\cite{zhou2020spatial} & - & 25.0 & -& -\\
\hline
\textbf{PiSLTRc-R} (ours) & 4.9/4.2 & \textbf{21.8} & 5.1/4.4 & \textbf{22.9} \\
\hline
\end{tabular}}
\label{slrfor2014t}
\end{table}

\begin{table}[t]
\renewcommand\arraystretch{1.3}
\centering
\smallskip
\caption{The evaluation results on sign2gloss task on PHOENIX-2014 dataset.}
\resizebox{0.48\textwidth}{!}{
\begin{tabular}{c|cc|cc}
\hline
\textit{sign2gloss} & \multicolumn{2}{c|}{DEV} & \multicolumn{2}{c}{TEST} \\
\hline
Model & del/ins & WER & del/ins & WER \\
\hline
DeepHand~\cite{Koller2016DeepHH} & 16.3/4.6 & 47.1 & 15.2/4.6 & 45.1 \\
DeepSign~\cite{Koller2016DeepSH} & 12.6/5.1 & 38.3 & 11.1/5.7 & 38.8 \\
SubUNets~\cite{Camgz2017SubUNetsEH} & 14.6/4.0 & 40.8 & 14.3/4.0 & 40.7 \\
Staged-Opt~\cite{Cui2017RecurrentCN} & 13.7/7.3 & 39.4 & 12.2/7.5 & 38.7 \\
Re-Sign~\cite{Koller2017ReSignRE} & - & 27.1 & - & 26.8 \\
DNF~\cite{Cui2019ADN} & 7.8/3.5 & 23.8 & 7.8/3.4 & 24.4 \\
CNN-LSTM-HMM~\cite{Koller2020WeaklySL} & - & 26.0 & - & 26.0 \\
FCN ~\cite{Cheng2020FullyCN} & - & 23.7 & - & 23.9 \\
STMC (RGB) ~\cite{zhou2020spatial} & - & 25.0 & -& -\\
SBD-RL~\cite{Wei2021SemanticBD} & 9.9/5.6 & 28.6 & 8.9/5.1 & 28.6 \\
\hline
SLTR-R(our implementation) & 8.9/4.2 & 24.5 & 9.0/4.3 & 24.6 \\
\hline
\textbf{PiSLTRc-R} (ours) & 8.1/3.4 & \textbf{23.4} & 7.6/3.3 & \textbf{23.2} \\
\hline
\end{tabular}}
\label{slrfor2014}
\end{table}

\subsubsection{Qualitative Analysis on SLR}
In Figure~\ref{testexample}, we show two examples with different methods on the SLR task. Equipped with proposed approaches, our PiSLTRc model learns accurate sign gesture recognition and thus achieving significant improvements. Furthermore, we find that the model trained based on CTC loss function tends to predict "peak" on the continuous gestures. And our proposed CPTcn model is adequate to alleviate this situation. As shown in Figure~\ref{testexample}, the recognition of adjacent frames in a contiguous region is more precise. 

\subsubsection{Qualitative Analysis on SLT}

In Table~\ref{exampleslt}, we show several examples with different models on the SLT task. Compared with the vanilla SLTR model~\cite{Camgz2020SignLT}, our proposed PiSLTRc produces target sentences with higher quality and accuracy.

Comparing the translation results of the first example as illustrated in Table~\ref{exampleslt}, we see that "vierzehn (fourteen)" is  mistranslated as "sieben (seven)" in SLRT model. However, it is correctly translated in our PiSLTRc model.  As we move to the second example in this table,  we see that "heute nacht (tonight)" is mistranslated as "morgen (tomorrow)" in SLRT model, and it is correctly in our PiSLTRc model.  To sum up, specific numbers and named entities are challenging since there is no grammatical context to distinguish one from another. However, in these two examples, we see that our model translates specific numbers and named entities more precisely. This demonstrates that our proposed model has a stronger ability to understand sign videos.

When we move to the third and fourth example in the Table~\ref{exampleslt}, we see that our model generate complete sentence with less under-translation. For example, in the third example, "gebietsweise länger andauernder regen (rain lasting longer in some areas)" is under-translated in SLTR model, while it is correctly translated as "bis zum teil dauerregen (partly constant rain)" in our PiSLTRc model.

In summary,  our proposed model performs better than the previous SLTR model when facing the specifical numbers and name entities, which are challenging  to translate since there is no grammatical context to distinguish one from another. Moreover, the sentences produced follow standard grammar. Nevertheless, it may be improved on the translation quality of the long sentences in the future.

\subsubsection{Limitation} We leverage neighboring similar features to enhance sign representation. The selected features are in a fixed-size region. This is not consistent with the characteristics of sign language. That is to say, the number of frames corresponding to different sign gestures is dynamic.


\begin{table*}[th]
\renewcommand\arraystretch{1.3}
\centering
\caption{The evaluation results on sign2text task on on Phoenix2014\textbf{T} dataset.}
\resizebox{\textwidth}{!}{
\begin{tabular}{l|ccccc|ccccc}
\hline
\textit{sign2text}& \multicolumn{5}{c|}{\textbf{DEV}} & \multicolumn{5}{c}{\textbf{TEST}} \\ 
\hline
Model & ROUGE & BLEU-1 & BLEU-2 & BLEU-3 & BLEU-4 & ROUGE & BLEU-1 & BLEU-2 & BLEU-3 & BLEU-4  \\
\hline
RNN-based\cite{Camgz2018NeuralSL} & 31.80 & 31.87 & 19.11 & 13.16 & 9.94 & 31.80 & 32.24 & 19.03 & 12.83 & 9.58  \\
TSPnet~\cite{Li2020TSPNetHF} & - & - & - & - & - & 34.96 & 36.10 & 23.12 & 16.88 & 13.41 \\
SLTR-T~\cite{Camgz2020SignLT} & - & 45.54 & 32.60 & 25.30  & 20.69 & -& 45.34 & 32.31 & 24.83 & 20.17 \\
Multi-channel~\cite{Camgz2020MultichannelTF} & 44.59 & - & - & - & 19.51 & 43.57 & - & - & - & 18.51 \\
\hline
PiSLTRc-T (ours) & \textbf{47.89} & \textbf{46.51} & \textbf{33.78} & \textbf{26.78} & \textbf{21.48} & \textbf{48.13} & \textbf{46.22} & \textbf{33.56} & \textbf{26.04} & \textbf{21.29} \\
\hline
\end{tabular}}
\label{sign2texttable}
\end{table*}

\begin{table*}[th]
\renewcommand\arraystretch{1.3}
\centering
\caption{The evaluation results on sign2(gloss+text) task on on Phoenix2014\textbf{T} dataset.}
\resizebox{\textwidth}{!}{
\begin{tabular}{l|c|ccccc|c|ccccc}
\hline
\textit{sign2(gloss+text)}& \multicolumn{6}{c|}{\textbf{DEV}} & \multicolumn{6}{c}{\textbf{TEST}} \\ 
\hline
Model & WER & ROUGE & BLEU-1 & BLEU-2 & BLEU-3 & BLEU-4 & WER & ROUGE & BLEU-1 & BLEU-2 & BLEU-3 & BLEU-4  \\
\hline
RNN-based\cite{Camgz2018NeuralSL} & - &  44.14 & 42.88 & 30.30 & 23.02 & 18.40 & - & 43.80 & 43.29 & 30.39 & 22.82 & 18.13 \\
SLTR~\cite{Camgz2020SignLT} & 24.61 & - & 46.56 & 34.03 & 26.83 & 22.12 & 24.49 & - & 47.20 & 34.46 & 26.75 & 21.80 \\
Multi-channel~\cite{Camgz2020MultichannelTF} & - & - & - & - & - & 22.38 & - & - & - & - & - & 21.32 \\
STMC(RGB-based)~\cite{9354538} & - & 44.30 & 44.06 & 32.69 & 25.45 & 20.74 & -& 44.70 & 45.08 & 33.80 & 26.44 & 21.55 \\
\hline
PiSLTRc (ours) & \textbf{22.23} & \textbf{49.87} & \textbf{47.37} & \textbf{35.41} & \textbf{28.09} & \textbf{23.17} & \textbf{23.01} & \textbf{49.72} & \textbf{48.50} & \textbf{35.97} & \textbf{28.37} & \textbf{23.40} \\
\hline
\end{tabular}}
\label{sign2glosstext}
\end{table*}

\begin{table}[t]
\renewcommand\arraystretch{1.2}
\centering
\smallskip
\caption{The evaluation results on sign2gloss task on CSL dataset.}
\resizebox{0.30\textwidth}{!}{
\begin{tabular}{c|c}
\hline
\textit{sign2gloss} & WER \\
\hline
S2VT~\cite{Venugopalan2015SequenceTS}  & 25.0 \\
LS-HAN~\cite{Huang2018VideobasedSL}  & 17.3 \\
HLSTM-attn~\cite{Guo2018HierarchicalLF}  & 10.2 \\
CTF~\cite{Wang2018ConnectionistTF}  & 11.2 \\
DenseTCN~\cite{Guo2019DenseTC} & 14.3  \\
SF-Net~\cite{Yang2019SFNetSF} & 3.8 \\
FCN~\cite{Cheng2020FullyCN} & 3.0 \\
\hline
SLTR-R(our implementation) &  3.7 \\
\hline
\textbf{PiSLTRc-R} (ours)  & \textbf{2.8} \\
\hline
\end{tabular}}
\label{slrforcsl}
\end{table}

\subsection{Comparison Against Baselines}
In this section, we compare several state-of-the-art models to demonstrate the effectiveness of our work. Similar to Camgz \textit{et al.}~\cite{Camgz2020SignLT}, we elaborate the comparison between our proposed model and baseline models in the three tasks: sign2gloss, sign2text, and sign2(gloss+text).

\subsubsection{sign2gloss} We evaluate this task in three datasets: PHOENIX-2014-T, PHOENIX-2014 and CSL.

In Table~\ref{slrfor2014t}, we compare our model with several methods for the \textit{sign2gloss} task on PHOENIX-2014-T dataset. DNF~\cite{Cui2019ADN} adopt iterative optimization approaches to tackle the weakly supervised problem. They first train an end-to-end recognition model for alignment proposal, and then use the alignment proposal to tune the feature extractor. CNN-LSTM-HMM~\cite{Koller2020WeaklySL} embeds powerful CNN-LSTM models in multi-stream HMMs and combines them with intermediate synchronization constraints among multiple streams. Vanilla SLTR-R~\cite{Camgz2020SignLT} uses the backbone pretrained with CNN-LSTM-HMM setup and then employes a two-layered transformer encoder model. FCN~\cite{Cheng2020FullyCN} is built upon an end-to-end fully convolutional neural network for cSLR. Furthermore, they introduce a Gloss Feature Enhancement (GFE) to enhance the frame-wise representation, where GFE is trained to provide a set of alignment proposals for the frame feature extractor. STMC (RGB)~\cite{zhou2020spatial} proposes a spatial-temporal multi-cue network to learn the video-based sequence. For a fair comparison, we only selected the RGB-based model of STMC without leveraging the additional information of hand, face, and body pose. PiSLTRc-R is our model which is trained when the weight of translation loss $\lambda_T$ is set zero. Similar to vanilla SLTR-R, our work extracts feature from the CNN-LSTM-HMM backbone. As shown in this table, our proposed PiSLTRc-R surpasses the vanilla SLTR model by $12\%$ and $7\%$ on Dev and Test datasets, respectively. Furthermore, in the RGB-based models, we achieve state-of-the-art performance on the \textit{sign2gloss} task. 

In Table~\ref{slrfor2014} we also evaluate our PiSLTRc-R model on the PHOENIX-2014 dataset. Compared with existed baseline models, our proposed model achieves comparable results. Note that the vanilla SLTR-R does not report the experimental results on the PHOENIX-2014 dataset. We implement it by ourselves. Compared with SLTR-R, our PiSLTRc-R model gains $4\%$ and $5\%$ improvements on Dev and Test datasets, respectively.

In Table~\ref{slrforcsl} we conduct experiments on CSL dataset. We see that our proposed PiSLTRc-R model achieves state-of-the-art performance. Compared with the SLTR-R model, our PiSLTRc-R model gains $24\%$ improvements on the Test datasets (5,000 examples split by ourselves), respectively.

\subsubsection{sign2text} In Table~\ref{sign2texttable}, we compare our approach with several \textit{sign2text} methods on PHOENIX-2014-T dataset. The RNN-based model~\cite{Camgz2018NeuralSL} adopt full frame features from Re-sign. TSPnet~\cite{Li2020TSPNetHF} utilizes I3D~\cite{Carreira2017QuoVA} to extract the spatial features, and further finetune I3D on two WSLR datasets~\cite{Li2020WordlevelDS,Joze2019MSASLAL}. Multi-channel~\cite{Camgz2020MultichannelTF} allows both the inter and intra contextual relationship between different asynchronous channels to be modelled within the transformer network itself. PiSLTRc-T is our model that training with the weight of recognition loss $\lambda_R$ being zero. Like in \textit{sign2gloss}, SLTR-T and our PiSLTRc-T model utilize the pretrained feature from CNN-LSTM-HMM. Experimental results show that our proposed model achieves state-of-the-art performance and surpasses the vanilla SLTR-T model by $3.8\%$ and $5.6\%$ BLEU-4 scores.

\subsubsection{sign2(gloss+text)} In Table~\ref{sign2glosstext}, we compare our model on \textit{sign2(gloss+text)} task. In this task, we jointly learn sign language recognition and translation simultaneously. Namely, $\lambda_R$ and $\lambda_T$ are set as non-zero. Note that different settings will obtain different results. Weighing up the performance on recognition and translation in our experiments, we set $\lambda_R=\lambda_T=1.0$. Compared with vanilla SLTR, our model gains significant improvements on both two tasks. Experiments demonstrate that our proposed techniques bring significant improvements for recognition and translation quality based on the sign language Transformer model. 

\section{Conclusion}
In this paper, we indicate two drawbacks of the sign language Transformer (SLTR) model for sign language recognition and translation. The first shortcoming is that self-attention aggregates sign visual features in a frame-wise manner, thus neglecting the temporal semantic structure of sign gestures. To overcome this problem, we propose a CPTcn module to generate neighborhood-enhanced sign features by leveraging the temporal semantic consistency of sign gestures. Specifically, we introduce a novel content-aware neighborhood gathering method to select relevant features dynamically. And then, we apply position-informed temporal convolution layers to aggregate them.

The second disadvantage is the absolute position encoding used in the vanilla SLTR model. It is demonstrated unable to capture the direction and distance information, which are critical for sign video understanding and sentence learning. Therefore, we inject relative position information to SLTR model with disentangled relative position encoding (DRPE) method. Extensive experiments on two large-scale sign language datasets demonstrate the effectiveness of our PiSLTRc framework.


%





\ifCLASSOPTIONcaptionsoff
  \newpage
\fi



\bibliographystyle{IEEEtran}
\bibliography{IEEEabrv,mylib}

\end{document}